%% file: main.tex
\documentclass[10pt,letterpaper]{article}
\usepackage{spconf,amsmath,amssymb,graphicx,booktabs,url}
\usepackage[hidelinks,bookmarks=false]{hyperref}
\input{frontmatter}
\begin{document}
\maketitle

\input{sections/abstract}
\input{sections/introduction}
\input{sections/method}
\input{sections/experimental_setup}
\input{sections/results}
\input{sections/conclusion}
\input{sections/disclosures}
\input{sections/references}
\end{document}

%% file: frontmatter.tex
\title{NuJEV: Structured Visual Judgments for Consistent Nuclei Quantification}
\name{Yonghuang Wu\sthanks{Equal contribution. $\dagger$Correspondence: \href{mailto:jhyu@fudan.edu.cn}{\texttt{jhyu@fudan.edu.cn}}.}, Pengfei Wang\textsuperscript{\ensuremath{\ast}}, Zhenyang Liang, Yize Ma, Pengfei Song, Guoqing Wu, Jinhua Yu\textsuperscript{\ensuremath{\dagger}}}
\address{Fudan University}

%% file: sections/abstract.tex
\begin{abstract}

Nuclear quantification requires locating and typing nuclei, beyond generating a count. Inspired by Jev's structured decisions in place of free-form generation, we introduce NuJEV for nuclear perception in Qwen3-VL-2B. We jointly train visual LoRA and a parallel head to predict nuclear presence, centers, and types. One image encoding feeds both the nuclear head and original language decoder. The decoder interprets region and type requests, with counts computed from the predicted nuclei. On 2,732 PanNuke test images, NuJEV achieves a five-query mean absolute error ($\mathrm{MAE}_5$) of 1.844 overall and 5.268 on dense images, with localization/typed F1 of 0.816/0.682. A matched encoder ablation shows 41.0\% lower dense error with visual adaptation. For language requests, NuJEV reduces request MAE by 76.2\% relative to direct generation. Our code will be released as open source.
\end{abstract}

%% file: sections/introduction.tex
\section{Introduction}

Quantitative histology requires counts by nuclear type and image region, rather than a single total. Language-based interaction lets analysts express these requests, but answering them requires accurate localization and typing. How can a vision--language model learn this fine-grained perception and use it for quantitative requests?

HoVer-Net~\cite{hovernet} and CellViT~\cite{cellvit} recover nuclei through segmentation; PointNu-Net~\cite{pointnu} and NuDetect~\cite{nudetect} detect them directly. Parallel point prediction~\cite{p2p} provides explicit locations, whereas coordinate generation~\cite{ptc} uses sequential decoding. We learn nuclear predictions in the dialogue model's visual encoder to share image features across perception and interaction.

We introduce NuJEV, extending Qwen3-VL-2B~\cite{qwen}. Inspired by Jev's direct structured decisions instead of free-form generation~\cite{jev}, we learn nuclear presence, position, and type through jointly trained visual adapters and a parallel head. The same image encoding feeds the original language decoder, which interprets region and type requests for counting from these predictions.

Six baselines and CoNSeP transfer assess nuclear perception; encoder ablations test visual adaptation. Matched language comparisons on all 2,732 PanNuke test images examine whether the predictions improve quantitative answers.

%% file: sections/method.tex
\section{Methods}
\begin{figure}[t]
\centering
\includegraphics[width=0.96\columnwidth,trim=29pt 46pt 30pt 41pt,clip]{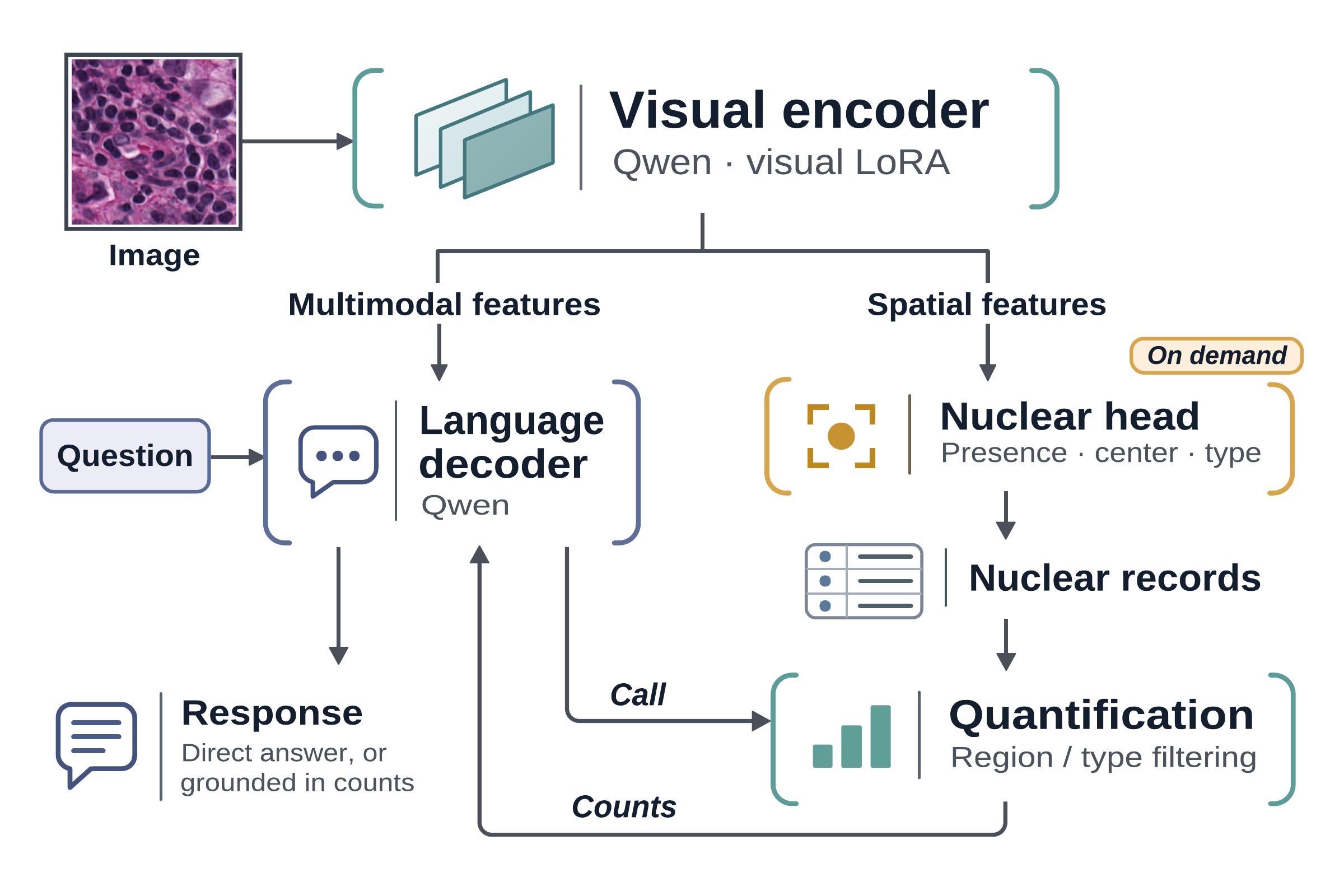}
\caption{
NuJEV jointly trains visual LoRA and a nuclear head. Shared image features also feed Qwen's language decoder. Counting operations use the predictions. Tissue: PanNuke development image.}
\label{fig:method}
\end{figure}

NuJEV shares image encoding between nuclear prediction and language interaction (Fig.~\ref{fig:method}). Its nuclear branch learns Jev-inspired structured judgments for quantitative requests.

\subsection{Learning Nuclear Perception}

Proposals encode centers and six-class probabilities: background, neoplastic, inflammatory, connective, dead, and epithelial. We train the head and rank-64 LoRA~\cite{lora} across 24 visual blocks, freezing pretrained weights. Inputs are resized from $256\times256$ to $512\times512$. Pre-merger $32\times32\times1024$ features pass through a 128-channel $1\times1$ projection and two $3\times3$ group-normalized GELU convolutions, producing four proposals per grid cell ($K=4096$):
\begin{equation}
 \hat p_k=a_k+4\Delta_k,\qquad \pi_k=\operatorname{softmax}(z_k)\in[0,1]^6.
\end{equation}

Here, $a_k$, $\Delta_k$, and $z_k$ denote anchor position, offset, and class logits. Hungarian matching~\cite{p2p} assigns targets by minimizing $0.25\|\hat p_k-p_j\|_2-\pi_{k,c_j}$; unmatched proposals are background. Training combines cross-entropy (background weight 0.1) and matched smooth-$L_1$ coordinate loss after dividing coordinates by 4, without truncating targets.

At decoding, presence $s_k=1-\pi_{k,0}$ and type derive from the same distribution; type is the highest-probability foreground class. We clip centers to image bounds, retain $s_k\geq0.60$, and apply score-ordered, class-independent 8-pixel suppression. Dense training images ($\geq60$ nuclei; 196/4,126) receive replacement-sampling weight 4 versus 1; rotations/reflections jointly transform images and centers.

\subsection{Language Integration}

Merged embeddings and DeepStack features feed the unchanged language decoder. Given image and question, Qwen answers directly or selects region, type, and output arguments. The head uses cached spatial features to predict centers and types $S(I)=\{(p_k,c_k)\}$. We select nuclei within half-open rectangle $R$ and class set $A$:
\begin{equation}
 \hat n(I,R,A)=\sum_{(p,c)\in S(I)}\mathbf{1}[p\in R]\mathbf{1}[c\in A].
\end{equation}

Counts (total, type, or quadrant) return to Qwen. Features and records are cached per request and cleared for new images; descriptions skip the head. Invalid arguments trigger corrective feedback. If the structured result is omitted, the interface appends the current successful tool output.

%% file: sections/experimental_setup.tex
\section{Experimental Setup}
\label{sec:baselines}
 
\textbf{Nuclear prediction.} PanNuke~\cite{pannuke} has 4,126 training, 1,043 development, and 2,732 test images with disjoint keys; patient and slide identifiers are unavailable.  Development selects checkpoints and decoding settings, fixed for test evaluation and ablations.  Dense images have $\geq60$ nuclei (50 development, 145 test); inference is identical across densities.

NuJEV uses a template parser without language decoding for fixed-query benchmarks. The nuclear branch is evaluated with 12 counting requests per image: $\mathrm{MAE}_5$ averages count errors over the full image, left/right halves, and two fixed random rectangles; seven further requests cover five full-image types and two regional types. Pooled localization F1 uses one-to-one 6-pixel matching; typed F1 also requires class agreement.  Point methods filter predictions by region and type; Qwen-Count answers requests directly.

\textbf{CellViT} and \textbf{HoVer-Net}~\cite{cellvit,hovernet} train from scratch with instance/pixel supervision in 100-epoch searches;  development selects dense-adapted CellViT and original HoVer-Net. \textbf{Qwen-Count/Points} adapt Qwen3-VL-2B with rank-32 language/visual LoRA (38.80M parameters), 1,284/1,200 updates, and greedy limits of 32/4,096 tokens.  Invalid counts and point lists are scored as zero and empty predictions, respectively. \textbf{CountGD} and \textbf{Grounding DINO}~\cite{countgd,gdino} adapt the last visual stage/detection components for 1,200 updates, batch eight, with density weighting/augmentation and frozen BERT.  CountGD uses $2\times2$ center boxes; Grounding DINO uses instance boxes. Each predicts six independently conditioned sets, without joint typed centers. 

 Sampling ablations match initialization, seed, decoder, 300 AdamW updates, and batch size 64: adapter/head rates $10^{-5}/2\times10^{-5}$, decay 0.01, warmup 20, cosine schedule, gradient clipping 1 (25.60M parameters).  
Both encoder arms start from the same original visual weights and trained head, with inherited adapters disabled. One trains only the head; the other also trains visual LoRA. Seed, sampling, augmentation, updates, batch, and decoder match; head and visual-LoRA rates are $10^{-4}$ and $5\times10^{-5}$.

\textbf{Language interaction.}  Each test image has a type-count and a quadrant-count request (5,464 per method).  
All three methods share images, the adapted visual encoder, and original language weights. Qwen-Direct answers from images; Qwen-Records also reads predicted centers and types; Ours accesses the trained nuclear head through counting operations. The latter two share identical full-precision predictions. Greedy generation allows 384 tokens/turn;  Ours allows four generation turns and two tool calls with feedback.  Request MAE averages absolute errors over requested counts, then tasks and images; invalid outputs use zero predictions.  We also report validity and exact agreement with independently counted predictions.  
We separately check non-counting descriptions and shared-encoding efficiency.

%% file: sections/results.tex
\section{Results}
\begin{figure*}[t]
\centering
\includegraphics[width=\textwidth]{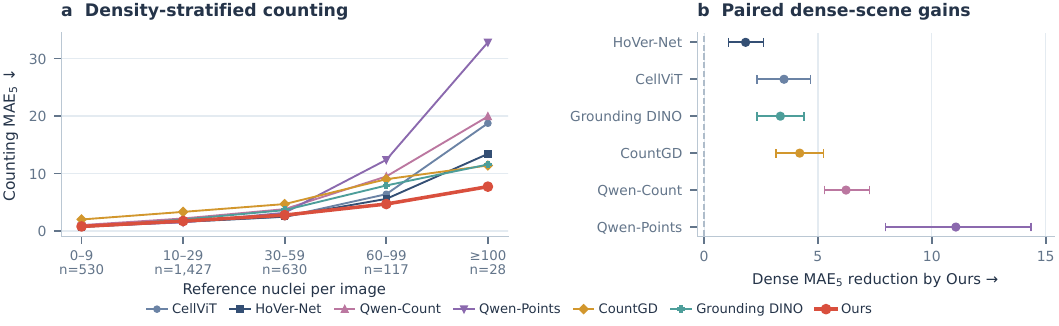}
\caption{ PanNuke test ($n=2,732$). (a) Five density groups. (b) Baseline-minus-Ours dense error; positive favors Ours. Intervals: 95\% paired image-bootstrap (2,000 resamples, $n=145$), excluding patient clustering and training-seed variation.}
\label{fig:density}
\end{figure*}
\begin{figure*}[t]
\centering
\includegraphics[width=\textwidth]{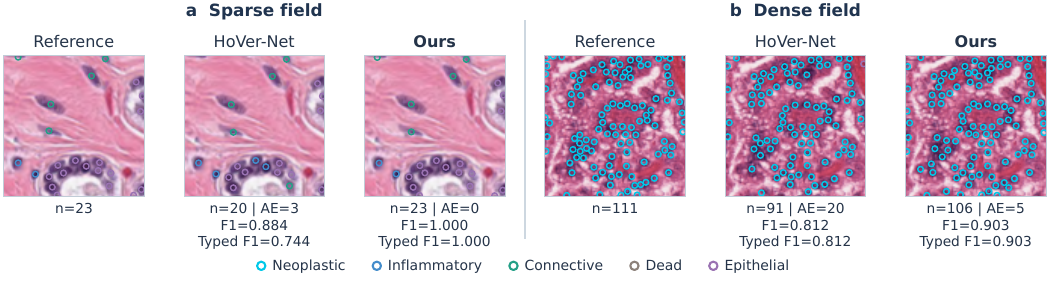}
\caption{ Sparse and dense examples selected for accurate Ours predictions and improvement over HoVer-Net in the dense case. $n$: count; $\mathrm{AE}=|n-n_{\mathrm{GT}}|$. F1: 6-pixel matching; typed F1 adds class agreement.  Test-set averages: Table~\ref{tab:final-test}.}
\label{fig:qualitative-main}
\end{figure*}
\input{tables/historical_results}
\input{tables/ablation}
\input{tables/consep_results}
\input{tables/interaction_results}

\textbf{Nuclear prediction.} On PanNuke, NuJEV's nuclear branch leads overall and dense-image counting and both F1 scores (Table~\ref{tab:final-test}), reducing MAE by \textbf{4.5\%} overall and \textbf{25.7\%} on dense images versus HoVer-Net. The paired dense-image difference (Ours minus HoVer-Net) is $-1.822$ (95\% interval $[-2.612,-1.068]$; Fig.~\ref{fig:density}). Non-dense error remains 1.652 versus CellViT's 1.606 and HoVer-Net's 1.642. Selected predictions appear in Fig.~\ref{fig:qualitative-main}.

\textbf{Visual adaptation.} With matched training, visual LoRA reduces MAE by \textbf{12.5\%} overall and \textbf{41.0\%} on dense images versus freezing, improving both F1 scores (Table~\ref{tab:training}b, one seed). Dense sampling and augmentation also reduce dense error (Table~\ref{tab:training}a). Paired 95\% intervals are $[-0.350,0.004]$ for dense versus uniform sampling and $[-0.556,0.110]$ for adding augmentation; both include zero.

\textbf{CoNSeP.}  Images ($1000\times1000$) use padded $4\times4$ tile grids. Out-of-image points are discarded; Qwen-Count queries only valid regions.  Ours reduces MAE by \textbf{30.2\%} relative to CellViT and achieves the highest F1 (Table~\ref{tab:consep}).

\textbf{Language interaction.} Using the learned nuclear predictions, NuJEV reduces request MAE by \textbf{76.2\%} and \textbf{80.8\%} versus direct and record-based answers (Table~\ref{tab:interaction}). Parameters are correct on 4,146/5,464 first calls and all final calls after feedback. All structured results match independent counts of predicted nuclei; request MAE also reflects nuclear prediction errors. Describing colors and nuclear morphology on 16 test images triggers neither the tool nor the head (0/16).

\textbf{Shared encoding.}  On eight test images (three paired repeats, MetaX C500), sharing reduces encoder passes from three to one and full-request time from 2.036 to 1.918\,s (5.8\%). Both peak at 4.74\,GiB allocated. Counts agree in 22/24 pairs; two differ by one nucleus.

%% file: tables/historical_results.tex
\begin{table}[t]
\centering
\caption{
Nuclear quantification on PanNuke: 2,732 test images (145 dense). $\mathrm{MAE}_5$ ($\downarrow$); pooled localization and typed F1 ($\uparrow$). Bold: best; ---: unavailable. Gains: relative MAE reduction or F1 increase versus the best baseline (+ better, $-$ worse).}
\label{tab:final-test}
\small
\setlength{\tabcolsep}{1.1pt}
\begin{tabular}{lrrrrr}
\toprule
Method & All & Dense & Non-dense & F1 & Typed F1\\
\midrule
CellViT &1.986&8.778&\textbf{1.606}&0.769&0.644\\
HoVer-Net &1.931&7.090&1.642&0.787&0.583\\
Qwen-Count &2.813&11.499&2.326&---&---\\
Qwen-Points &2.529&16.321&1.755&0.677&0.481\\
CountGD &3.698&9.466&3.374&0.727&---\\
Grounding DINO &2.465&8.617&2.120&0.777&---\\
\midrule
\textbf{Ours} &\shortstack{\textbf{1.844}\\(+4.5\%)}&\shortstack{\textbf{5.268}\\(+25.7\%)}&\shortstack{1.652\\(-2.9\%)}&\shortstack{\textbf{0.816}\\(+3.6\%)}&\shortstack{\textbf{0.682}\\(+5.8\%)}\\
\bottomrule
\end{tabular}
\end{table}

%% file: tables/ablation.tex
\begin{table}[t]
\centering
\caption{
PanNuke test ablations: initialization, seed, decoder, and 300 updates are matched within each panel. $\mathrm{MAE}_5$ ($\downarrow$); F1 ($\uparrow$). Bold: best per panel. Gains versus (a) the best other row per column; (b) the frozen encoder.}
\label{tab:training}
\small
\setlength{\tabcolsep}{2.2pt}
\begin{tabular}{lrrrr}
\toprule
\multicolumn{5}{l}{(a) Sampling and augmentation}\\
Training & All & Dense & Non-dense & Dense F1\\
\midrule
Uniform &1.896&5.650&1.686&0.8347\\
Dense sampling &1.900&5.477&1.700&0.8353\\
\textbf{Ours} &\shortstack{\textbf{1.844}\\(+2.8\%)}&\shortstack{\textbf{5.266}\\(+3.9\%)}&\shortstack{\textbf{1.652}\\(+2.0\%)}&\shortstack{\textbf{0.8417}\\(+0.8\%)}\\
\midrule
\multicolumn{5}{l}{(b) Encoder adaptation}\\
Setting & All & Dense & F1 & Typed F1\\
\midrule
Frozen encoder &2.280&9.034&0.7496&0.5875\\
Visual LoRA &\shortstack{\textbf{1.994}\\(+12.5\%)}&\shortstack{\textbf{5.334}\\(+41.0\%)}&\shortstack{\textbf{0.8112}\\(+8.2\%)}&\shortstack{\textbf{0.6686}\\(+13.8\%)}\\
\bottomrule
\end{tabular}
\end{table}

%% file: tables/consep_results.tex
\begin{table}[t]
\centering
\caption{
CoNSeP test: 14 images (8,777 nuclei), 16 tiles per image; PanNuke-trained models evaluated without retraining. MAE/RMSE ($\downarrow$); 6-pixel F1 ($\uparrow$). Bold and gains as in Table~\ref{tab:final-test}.}
\label{tab:consep}
\small
\setlength{\tabcolsep}{3pt}
\begin{tabular}{lrrr}
\toprule
Method & MAE & RMSE & F1\\
\midrule
CellViT & 114.571 & 165.260 & 0.665\\
HoVer-Net & 123.643 & 179.803 & 0.688\\
Qwen-Count & 212.429 & 275.658 & ---\\
Qwen-Points & 190.571 & 246.909 & 0.560\\
CountGD & 151.929 & 203.113 & 0.588\\
Grounding DINO & 255.643 & 301.053 & 0.611\\
\midrule
\textbf{Ours} & \shortstack{\textbf{80.000}\\(+30.2\%)} & \shortstack{\textbf{110.849}\\(+32.9\%)} & \shortstack{\textbf{0.696}\\(+1.3\%)}\\
\bottomrule
\end{tabular}
\end{table}

%% file: tables/interaction_results.tex
\begin{table}[!tb]
\centering
\caption{
PanNuke test: 2,732 images, 5,464 type/quadrant requests/method. MAE: ground-truth error. Valid: parseable required integer counts. Agreement: exact match to independent counts of predicted nuclei. Gains: relative MAE reduction or percentage-point (pp) increases versus best other row.}
\label{tab:interaction}
\small
\setlength{\tabcolsep}{3pt}
\begin{tabular}{lrrr}
\toprule
Method & MAE $\downarrow$ & \shortstack[r]{Valid\\output $\uparrow$} & \shortstack[r]{Prediction\\agreement $\uparrow$}\\
\midrule
Qwen-Direct &5.991&5359/5464&---\\
Qwen-Records &7.420&5441/5464&195/5464\\
\textbf{Ours} &\shortstack{\textbf{1.423}\\(+76.2\%)}&\shortstack{\textbf{5464/5464}\\(+0.4 pp)}&\shortstack{\textbf{5464/5464}\\(+96.4 pp)}\\
\bottomrule
\end{tabular}
\end{table}

%% file: sections/conclusion.tex
\section{Discussion and Conclusion}

NuJEV extends Qwen3-VL-2B with learned nuclear localization and typing. Joint visual adaptation and head training provide the predictions, while shared encoding connects them to the pretrained language decoder. Visual adaptation reduces dense error, and using nuclear predictions improves numerical responses.
\label{technical-end}

%% file: sections/disclosures.tex
\section{Compliance with Ethical Standards}
\label{sec:ethics}
We analyze publicly available PanNuke and CoNSeP images and collect no new human-subject data.

\section{Acknowledgments}
\label{sec:acknowledgments}
No funding was received for conducting this study. The authors have no relevant financial or non-financial interests to disclose.
We used OpenAI Codex to assist with literature exploration and code inspection. We retain responsibility for the scientific content.

%% file: sections/references.tex
\bibliographystyle{IEEEbib-initials}
\bibliography{references}